\documentclass[lettersize,journal]{IEEEtran}
\usepackage{amsmath,amsfonts}
\usepackage{algorithmic}
\usepackage{algorithm}
\usepackage{array}
\usepackage[caption=false,font=normalsize,labelfont=sf,textfont=sf]{subfig}
\usepackage{textcomp}
\usepackage{stfloats}
\usepackage{url}
\usepackage{verbatim}
\usepackage{graphicx}
\usepackage{cite}
\usepackage{booktabs}
\usepackage{multirow}
\def\BibTeX{{\rm B\kern-.05em{\sc i\kern-.025em b}\kern-.08em
    T\kern-.1667em\lower.7ex\hbox{E}\kern-.125emX}}

\begin{document}

\title{Lightweight Structure-aware Transformer Network for VHR Remote Sensing Image Change Detection}
\author{Tao Lei, ~\IEEEmembership{Senior Member,~IEEE}, Yetong Xu, Hailong Ning, Zhiyong Lv, Chongdan Min, Yaochu Jin, ~\IEEEmembership{Fellow,~IEEE,} and Asoke K. Nandi, ~\IEEEmembership{Life Fellow,~IEEE}\vspace{-4.3mm}
\thanks{This work was supported in part by National Natural Science Foundation of China (Program No. 62271296, 62201452), in part by the Natural Science Basic Research Program of Shaanxi (Program No. 2021JC-47, 2022JQ-592), in part by Key Research and Development Program of Shaanxi (Program No. 2022GY-436, 2021ZDLGY08-07), in part by Shaanxi Joint Laboratory of Artificial Intelligence (Program No. 2020SS-03), and in part by Scientific Research Program Funded by Shaanxi Provincial Education Department (Program No. 22JK0568). (Corre-sponding author: Hailong Ning.)

T. Lei, Y. Xu and C. Min are with the Shaanxi Joint Laboratory of Artificial Intelligence, Shaanxi University of Science and Technology, Xi’an 710021,China. (e-mail: leitao@sust.edu.cn; xuyetong1999@163.com; bs221611001@sust.edu.cn).

H. Ning is with the School of Computer Science and Technology, Xi’an University of Posts and Telecommunications, Xi’an 710121, China. (e-mail: ninghailong93@gmail.com).

Z. Lv is with the School of Computer Science and Engineering, Xi’an University of Technology, Xi’an 710048, China. (E-mail: Lvzhiy-ong fly@hotmail.com).

Y. Jin is with the Faculty of Technology, Bielefeld University, 33619 Bielefeld, Germany (email: yaochu.jin@uni-bielefeld.de).

A. K. Nandi is with the Department of Electronic and Electrical Engi-neering, Brunel University London, Uxbridge, Middlesex, UB8 3PH, U.K.,
and visiting professor with Xi’an Jiaotong University, Xi’an 710049, China.
(e-mail: asoke.nandi@brunel.ac.uk).}}

\markboth{}%
{Shell \MakeLowercase{\textit{et al.}}: A Sample Article Using IEEEtran.cls for IEEE Journals}

\maketitle
\begin{abstract}
Popular Transformer networks have been successfully applied to remote sensing (RS) image change detection (CD) identifications and achieve better results than most convolutional neural networks (CNNs), but they still suffer from two main problems. First, the computational complexity of the Transformer grows quadratically with the increase of image spatial resolution, which is unfavorable to very high-resolution (VHR) RS images. Second, these popular Transformer networks tend to ignore the importance of fine-grained features, which results in poor edge integrity and internal tightness for largely changed objects and leads to the loss of small changed objects. To address the above issues, this Letter proposes a Lightweight Structure-aware Transformer (LSAT) network for RS image CD. The proposed LSAT has two advantages. First, a Cross-dimension Interactive Self-attention (CISA) module with linear complexity is designed to replace the vanilla self-attention in visual Transformer, which effectively reduces the computational complexity while improving the feature representation ability of the proposed LSAT. Second, a Structure-aware Enhancement Module (SAEM) is designed to enhance difference features and edge detail information, which can achieve double enhancement by difference refinement and detail aggregation so as to obtain fine-grained features of bi-temporal RS images. Experimental results show that the proposed LSAT achieves significant improvement in detection accuracy and offers a better tradeoff between accuracy and computational costs than most state-of-the-art CD methods for VHR RS images. 
\end{abstract}


\vspace{-2.5mm}\section{Introduction}
\IEEEPARstart{R}{emote} sensing (RS) image change detection (CD) aims to obtain information on surface changes in different periods but the same geographical area [1]. It has been widely applied in urban sprawl monitoring [2], land cover monitoring [3], and disaster assessment [4]. Currently, CD has become a significant research direction in the field of RS. 

In recent years, deep learning techniques based on CNNs have shown excellent performance in VHR RS image CD tasks [5,6]. For example, Daudt \emph{et al}. [7] first applied Siamese full convolutional networks (FCNs) to the CD and proposed three end-to-end networks, which established the mainstream framework for subsequent CD. Due to the complexity of RS image scenes and susceptibility to light and environment changes, various multi-scale feature fusion modules and attention mechanisms have been introduced into the Siamese network [8, 9, 10], which achieves satisfactory results. However, although CNN-based methods are effective in extracting discriminative features for CD, they struggle to model long-range contextual information in bi-temporal RS images.

Recently, Vision Transformer (ViT) has been successfully applied to CD tasks due to its excellent ability of capture long-range dependency relationships. For instance, Chen \emph{et al}. [11] proposed a bi-temporal image transformer (BIT) method to model the long-range contextual information in bi-temporal images. Different from the original ViT model, BIT uses the semantic token to represent the input image features, and thus it has fewer parameters and lower computational costs. Although BIT achieves good change detection results for VHR RS images, it is not a pure Transformer network due to the employment of ConvNets in its encoder. To solve the problem, ChangeFormer [12] drops the ConvNets encoder in BIT and only uses a Transformer encoder and a lightweight MLP decoder. Consequently, ChnageFormer provides better CD results than BIT. Unlike the above work, SwinSUNet [13] proposes a pure Swin transformer [14] network with Siamese U-shaped structure, which also achieves good results for VHR RS image CD. The above methods perform well in CD tasks but still suffer from two major challenges. First, the computational complexity of most Transformer-based CD methods grows quadratically with the increase of image spatial resolution, making it difficult to train a Transformer network for VHR RS image CD. Second, existing Transformer-based CD methods (e.g., BIT and ChangeFormer) ignore the importance of fine-grained information, resulting in suboptimal edge integrity and internal tightness for largely changed objects and the missed detection for small changed objects.

To tackle the above challenges, this Letter proposes a Lightweight Structure-aware Transformer (LSAT) method for VHR CD. Our LSAT is a U-shaped structure consisting of a dual-branch weight-sharing encoder and a single-branch decoder. Both the encoder and decoder are composed of Cross-dimension Interactive Self-attention (CISA) module with linear complexity. To integrate detailed change information from the dual-branch at each level into the corresponding decoding layer, we propose a Structure-aware Enhancement  Module (SAEM) between the encoder and the decoder. In addition, an Attention-based Fusion Module (AFM) is added after the encoding layer to fuse efficiently the bi-temporal deep semantic features. 
\begin{figure*}
\vspace{-3mm}
\setlength{\abovecaptionskip}{-0.15cm}
\setlength{\belowcaptionskip}{-0.0cm}
	\begin{center}
		\includegraphics[width=\linewidth]{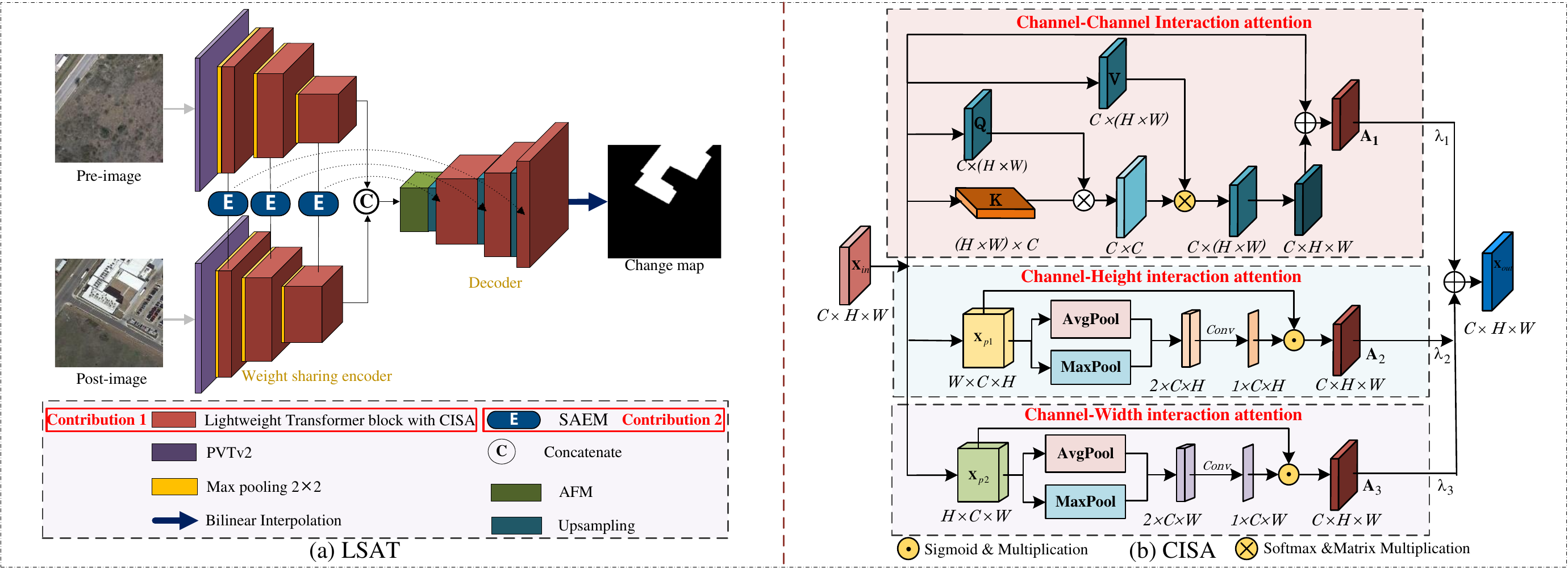}
		\renewcommand{\figurename}{Fig.}
	\end{center}
	\caption{\small{Overall network architecture. (a) LSAT architecture. (b) The specific architecture of the proposed CISA. The encoder is composed of a lightweight Transformer block using CISA. The CISA is composed of three branches of channel interaction attention, the first channel ${{\bf{A}}_{\bf{1}}}$ is the Channel-Channel attention branch, the second channel ${{\bf{A}}_{\bf{2}}}$ is the Channel-Height interaction attention branch, and the third channel ${{\bf{A}}_{\bf{3}}}$ is the Channel-Width interaction attention branch.}} \vspace{-3mm}
	\label{fig:1}
\end{figure*}

The main contributions of this Letter are summarized as follows:

\begin{enumerate}
\item{A Cross-dimension Interactive Self-attention (CISA) module with linear complexity is proposed to achieve a lightweight Transformer network. Different from the vanilla self-attention module, our CISA not only effectively reduces the computational complexity of networks, but also achieves excellent change detection accuracy for VHR RS images CD.}

\item{A Structure-aware Enhancement Module (SAEM) is designed to learn fine-grained features and improve the internal tightness of the largely changed objects. Different from regular single-branch difference enhancement methods, SAEM achieves a double enhancement by calibrating the difference and co-detail features with a dual-branch architecture.} 

\item{An efficient Lightweight Structure-aware Transformer (LSAT) network is proposed based on the employment of both CISA and SAEM. Extensive experiments on two typical datasets in CD demonstrate that the proposed LSAT consistently outperforms other state-of-the-art (SOTA) networks in detection accuracy and computational costs.}
\end{enumerate}

\section{METHODOLOGY}
The framework of the proposed LSAT includes three main modules as shown in Fig.1(a). First, the encoder with the CISA module is mainly used to extract hierarchical semantic features of bi-temporal RS images, and the SAEM module is used for enhancing the fine-grained difference features. Second, the attention-based fusion module (AFM) is mainly used to generate fine-grained change maps. Third, the fine-grained difference features and bi-temporal deep semantic features are integrated into the decoder to output changed objects in VHR RS images.

\subsection{Cross-dimension Interactive Self-attention}

In vanilla self-attention (SA), the computational complexity of the \textit{key-query} dot-product interaction grows quadratically with the increasing spatial resolution of input images, as $ O({(WH)^2})$ for images of size $W \times H$. This quadratic complexity greatly increases the training difficulty of the network for the VHR CD task. To solve the problem, we design a Cross-channel Interactive Self-attention (CISA) module with linear complexity as an alternative to the vanilla SA in the proposed Transformer architecture. The architecture of CISA module is illustrated in Fig. 1(b).

First, inspired by CvT [15], \textit{query} (${\bf{Q}}$), \textit{key} (${\bf{K}}$) and \textit{value} (${\bf{V}}$) are generated by depth-wise separable convolutional projections but not the linear projection, which not only strengthens the connection of local contexts, but also reduces semantic ambiguity caused by the vanilla self-attention mechanism.

Next, the inter-channel encoding is performed to generate a Channel-Channel attention map ${{\bf{A}}_{\bf{1}}}$. In addition, to capture cross-channel long-dependency relationships, the two-branch interaction attentions of Channel-Height ${{\bf{A}}_{\bf{2}}}$ and Channel-Width ${{\bf{A}}_{\bf{3}}}$  are conducted on features maps to enhance the cross-dimension interactions between channel and spatial dimensions. It can improve the global information extraction ability of the model. Each of the three types of attention can be expressed as:\vspace{-1mm}
\begin{equation}\label{Eq.:1}
	{{\bf{A}}_{{1}}}({\bf{Q,K,V}}) = {\mathop{\rm Softmax}\nolimits} (\frac{{{\bf{Q}}{{\bf{K}}^{\bf{T}}}}}{a}){\bf{V}} \vspace{-1.5mm},
\end{equation}
\begin{equation}\label{Eq.:2}
	{{\bf{A}}_{{2}}}({C, H}) = {\mathop{\rm Sigmoid}\nolimits} ({{C}_1} ({P} ({{\bf{X}}_{p1}}))){{\bf{X}}_{p1}} \vspace{-1.5mm},
\end{equation}
\begin{equation}\label{Eq.:3}
	{{\bf{A}}_{{3}}}({C, W}) = {\mathop{\rm Sigmoid}\nolimits} ({{C}_1} ({P} ({{\bf{X}}_{p2}}))){{\bf{X}}_{p2}},
\end{equation}
where ${{\bf{X}}_{p1}} \in {^{W \times C \times H}}$ and ${{\bf{X}}_{p}} \in {^{W \times C \times H}}$ are the tensors after performing the dimensional transformation for the input feature ${{\bf{X}}_{in}}$, respectively. ${{C}_1}$ is a $1 \times 1$ convolution operation, ${P}$ represents the parallel operation of maximum pooling and average pooling. The final attention is computed as follows: 
 \vspace{-1mm}
\begin{equation}\label{Eq.:4}
	{\bf{A}} = {\lambda _1}{{\bf{A}}_{{1}}} + {\lambda _2}{{\bf{A}}_{{2}}} + {\lambda _3}{{\bf{A}}_{{3}}}\vspace{-0.25mm},
\end{equation}
where ${\lambda _1}$, ${\lambda _2}$ and ${\lambda _3}$ are the pre-defined hyperparameters. 

Note that the  cross-channel self-attention in the ${{\bf{A}}_{{1}}}$ branch of the ${\bf{Q}}$, ${{\bf{K}}^{{T}}}$ dot-product interaction generates a transposed attention map of size ${R^{C \times C}}$ instead of the conventional attention map of size ${R^{HW \times HW}}$. In addition, the other two branches are mainly convolution operations, which have low computational complexity. Therefore, the computational complexity of CISA is $O({C^2+ CH + CW})$ much smaller than the conventional quadratic complexity $O({(WH)^2})$. To verify the effectiveness of the proposed CISA module, we present the experiments in Section III.
\subsection{Structure-aware Enhancement Module}
\begin{figure}[tp]
\vspace{-3mm}
\setlength{\abovecaptionskip}{-0.25cm}
\setlength{\belowcaptionskip}{-0.cm}
	\begin{center}
		\includegraphics[width=\linewidth]{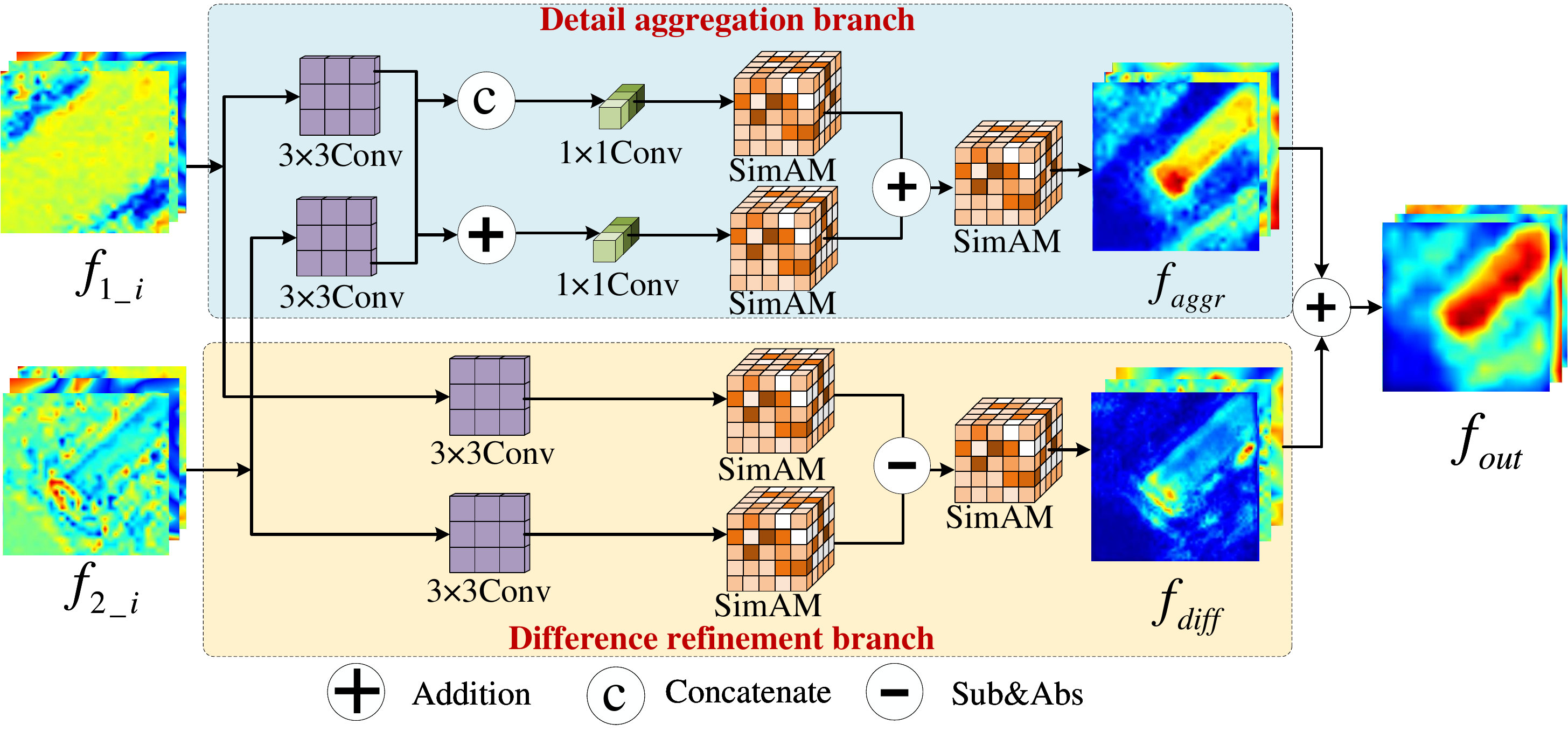}
		\renewcommand{\figurename}{Fig.}
		\vspace{-5 mm}
	\end{center}
	\caption{\small{SAEM structure, ${f_{1\_i}}$, ${f_{2\_i}}$ are the dual-time phase output features of each layer of the encoder. ${f_{diff}}$ is the enhanced feature of different-detail branch and ${f_{aggr}}$ is the enhanced feature of detail aggregation branch.}} \vspace{-5mm}
	\label{fig:2}
\end{figure}
In existing CD networks, the importance of fine-grained features is often ignored, resulting in poor edge integrity of objects with large-size changes and undetectable edge integrity of objects with small-size changes. To solve this problem, the difference image (DI) is often used to enhance the details of changed objects. However, single-branch difference enhancement methods, such as performing subtraction operation on bi-temporal images or employing an attention mechanism to improve feature extraction of bi-temporal images [5, 16]. This enhancement method is suboptimal because of the limited feature expression ability of single-branch enhancement and a high computational cost of the attention. Therefore, we propose a lightweight Structure-aware Enhancement Module (SAEM) to learn the difference information about CDs completely. Different from single-branch difference enhancement methods, SAEM performs double enhancement by using dual-branch to learn fine-grained features. Fig. 2. shows the details of SAEM.

Our SAEM consists of two branches: the difference refinement and the detail aggregation. In the difference refinement branch, the bi-temporal features are further enhanced by using convolution operation, and then a lightweight 3D attention SimAM [17] is used to generate finer-grained features and improve the contours of detection results. This process can be expressed as follows:
\begin{equation}\label{Eq.:5}
	{f_{diff}} = {{M}_a}\left| {{{M}_a}({{C}_3}({f_{1\_i}})) - {{M}_a}({{C}_3}({f_{2\_i}}))} \right|,
\end{equation}
where ${f_{1\_i}}$, ${f_{2\_i}}$ are the pre-temporal and post-temporal features at \emph{i}-th level respectively, ${{C}_3}$ is a $3 \times 3$ convolution operation, ${{M}_a}$ is a lightweight attention mechanism, and $\left|  \cdot  \right|$ denotes the absolute value operation to ensure the availability of obtained difference features.

The detail aggregation branch is subdivided into two pathways, one for enhancing the detail information by adding up the convolved features. The other pathway is to concatenate the convolution features and then use attention to extract richer detailed information. The two branches are computed as follows: \vspace{-1.5mm} 
\begin{equation}\label{Eq.:6}
	{f_{a1}} = {{C}_1}({{C}_3}({f_{1\_i}}) \oplus {{C}_3}({f_{2\_i}})),
\end{equation}
\begin{equation}\label{Eq.:7}
	{f_{a2}} = {{M}_a}[{{C}_3}({f_{1\_i}});{{C}_3}({f_{2\_i}})],
\end{equation}
where [;] is the concatenation operation. Finally, after dimensionality reduction, the attention is used to aggregate the detailed information of the two pathways, and it is defined as:\vspace{-1mm} 
\begin{equation}\label{Eq.:8}
	{f_{aggr}} = {{M}_a}({f_{a1}} \oplus {{M}_a}({f_{a2}})).
\end{equation}

The finally aggregated features contain richer edge details than that generated by a single-branch module of the bi-temporal RS images. The SAEM can reduce the erroneous changes caused by noise and misalignment, thus obtaining more useful fine-grained change features and increasing the robustness of our proposed network. The final output is calculated as follows: 
\begin{equation}\label{Eq.:9}
	{f_{out}} = {f_{diff}} \oplus {f_{aggr}}.\vspace{-1.5mm}
\end{equation}

To verify the effectiveness of the proposed SAEM, the experiments in Section III.
\section{EXPERIMENTS}

\subsection{Datasets and Evaluation Metrics}
In this Letter, experiments are conducted on two publicly available large CD datasets, LEVIR-CD [10] and CDD [18].

\textbf{LEVIR-CD dataset} contains 637 VHR Google Earth image pairs with a resolution of 0.5m, and with a size of $1024 \times 1024$. To prevent overfitting, data enhancement operations such as random rotation and random cropping are performed in this Letter, and the images are randomly cropped into patches with size of $256 \times 256$. 10,000 pairs are set aside for training, 1,024 pairs for validation, and 2,048 pairs for testing. 

\textbf{CDD dataset} is a RS image with seasonal variation of the same region acquired by Google Earth, and a total of 16,000 pairs of image pairs with a size of $256 \times 256$ are obtained through random cropping and data enhancement, of which 10,000 pairs are used for training, 3,000 pairs for validation, and the remaining 3,000 pairs for testing.

In this Letter, three main evaluation metrics, including Precision (\emph{Pre}), Recall (\emph{Rec}), F1-Score (\emph{F1}) and Distance from the ideal position (\emph{DIP}) [19] are used to evaluate comprehensively the network, where \emph{F1} and \emph{DIP} comprehensively consider the two metrics \emph{Pre} and \emph{Rec} in different ways.\vspace{-3mm}

\subsection{Implementation Details}
In this Letter, the proposed method is implemented by using PyTorch and trained for 200 epochs on an NVIDIA GeForce RTX 3090 GPU. The training process is performed using the AdamW optimizer and the momentum is set to 0.99. The weight decay is set to 0.0005 and the initial learning rate was 0.0001. Experiments show that 2:1:1 is the best value for ${\lambda _1}$ : ${\lambda _2}$ : ${\lambda _3}$, and the default value for subsequent experiments is 2:1:1. The pretrained PVTv2-B1 [20] is used for preliminary feature extraction. To mitigate the effect of the category imbalance problem, this Letter combines the binary cross-entropy loss with the Dice loss for optimizing the model.\vspace{-2mm}
\subsection{Comparison with State-of-the-art}
\subsubsection{\textbf{Comparison methods}}To demonstrate the effectiveness of our proposed LSAT, some popular SOTA networks are used for comparison, including FCN-PP [4], IFNet [5], STANet [10],  FDCNN [8], SNUNet [6], DSAMNet [9], BIT [11], ChangeFormer [12], and SwinSUNet [13].
\subsubsection{\textbf{Comparison on the LEVIR-CD dataset}}
The quantitative evaluation results of the LEVIR-CD dataset are shown in Table I. The best values are shown in bold for all the following tables. It can be seen that the proposed LSAT almost obtains the best results. Compared to ChangeFormer, our network achieves 1.12\%/1.24\% higher in \emph{F1} and \emph{DIP}, respectively. To further illustrate the superiority of LSAT, a visual analysis is conducted as shown in Fig. 3(a). Although most of the comparison methods show missed detection and false multiple detections, the proposed LSAT method has better detection results. It can be observed that our LSAT achieves the best performance in the integrity and accuracy of changed objects.
\begin{table}[htbp]
\vspace{-4mm}
\renewcommand\arraystretch{1.20}
 \tabcolsep=0.15cm
\renewcommand\arraystretch{1.1}
	\centering
	\caption{Performance comparison on the LEVIR-CD test set.}
\begin{tabular}{cccccc}
\hline
Method Type                  & Network              & \emph{Pre}(\%) &\emph{Rec}(\%) &\emph{F1}(\%)  &\emph{DIP}(\%) \\\hline
\multirow{6}{*}{CNN}         & FCN-PP {[}4{]}        & 80.31    & 89.48    & 84.64   & 84.21   \\
                             & STANet {[}10{]}       & 86.14    & 89.39    & 87.73   & 87.65   \\
                             & IFNet {[}5{]}          & 89.73    & 86.06    & 87.80   & 87.75   \\
                             & FDCNN {[}8{]}         & 82.99    & 88.71    & 85.76   & 85.56   \\
                             & SNUNet {[}6{]}        & 89.06    & 87.53    & 88.29   & 88.27   \\
                             & DSAMNet {[}9{]}       & 82.75    & 88.39    & 85.48   & 85.29   \\\hline
\multirow{3}{*}{Transformer} & BIT {[}11{]}          & 89.24    & 89.37    & 89.31   & 89.30   \\
                             & ChangeFormer {[}12{]} & \textbf{92.05}    & 88.80    & 90.40   & 90.28   \\
                             & SwinSunet {[}13{]}    & 90.51    & 89.72    & 90.11   & 90.10   \\
                             & \textbf{LSAT(ours)}  & 91.81    & \textbf{91.24}    & \textbf{91.52}  & \textbf{91.52}\\\hline
\end{tabular}

	\label{tabl}%
\end{table}%
\subsubsection{\textbf{Comparison on the CDD dataset}}
In Table II, the proposed LSAT also obtains the best results on the CDD dataset, compared to SwinSUNet, our LSAT achieves 1.12\%/2.04\% higher in \emph{F1} and \emph{DIP}, respectively. A clear visual analysis is shown in Fig. 3(b). For the complex change region detection, the comparative networks provide detection results showing leakages and inaccuracy on contours, while our LSAT still provides the best detection contours.

All the above experimental results show that our LSAT can effectively improve the performance of VHR CD since both long-range contextual information and fine-grained information are effectively captured by the proposed LSAT. In this way, we can enhance the edge integrity and internal tightness for changed objects, and reduce the missed detection for small changed objects. Furthermore, we conducted more experiments in our supplement materials.
\begin{table}[htbp]
\vspace{-3mm}
\renewcommand\arraystretch{1.20}
 \tabcolsep=0.15cm
\renewcommand\arraystretch{1.1}
	\centering
	\caption{Performance comparison on the CDD test set.}
\begin{tabular}{cccccc}
\hline
Method Type                  & Network             & \emph{Pre}(\%) &\emph{Rec}(\%) &\emph{F1}(\%)  &\emph{DIP}(\%) \\\hline
\multirow{6}{*}{CNN}         & FCN-PP {[}4{]}       & 81.69    & 90.31    & 85.78  & 85.35   \\
                             & STANet {[}10{]}      & 88.98    & 93.11    & 91.00  & 90.80   \\
                             & IFNet {[}5{]}         & 90.72    & 86.50    & 88.56  & 88.41   \\
                             & FDCNN {[}8{]}        & 83.61    & 91.70    & 87.47  & 87.01   \\
                             & SNUNet {[}6{]}       & 90.92    & 94.75    & 92.79  & 92.58   \\
                             & DSAMNet {[}9{]}      & 91.67    & 94.83    & 93.22  & 93.06   \\\hline
\multirow{3}{*}{Transformer} & BIT {[}11{]}        & 92.89    & 94.02    & 93.45  & 93.43   \\
                             & ChangeFormer {[}12{]}& 94.26    & 93.46    & 93.84  & 93.84   \\
                             & SwinSUNet {[}13{]}   & 95.70    & 92.30    & 94.00  & 93.76   \\
                             & \textbf{LSAT(ours)} & \textbf{97.02}    & \textbf{94.87}    & \textbf{95.93}  & \textbf{95.80} \\\hline
\end{tabular}\vspace{-3mm}

	\label{tab2}%
\end{table}%

\begin{figure*}
\vspace{-4mm}
\setlength{\abovecaptionskip}{-0.3cm}
\setlength{\belowcaptionskip}{-0.2cm}
	\begin{center}
		\includegraphics[width=\linewidth]{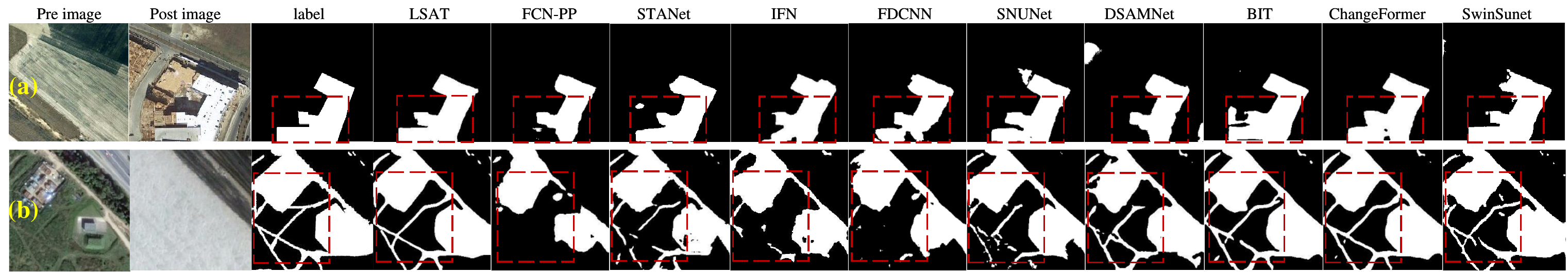}
		\renewcommand{\figurename}{Fig.}
	\end{center}\caption{\small{ Visual analysis of the detection results, (a)LEVIR-CD, (b)CDD.}}
	\label{fig:3}
\end{figure*}

\subsection{Ablation Study}
To verify the effectiveness of the proposed modules, we performed ablation experiments on the LEVIR-CD for CISA, SAEM, and AFM modules, respectively. The specific results are shown in Table III. Siamese-unet is considered as the baseline. From the experimental results, we can see that the proposed CISA not only improves the \emph{F1} score by nearly 2.79\% over the baseline, but also significantly reduces the computational complexity. When adding SAEM to CISA for adequately capturing and fusing the multi-scale detail information of changed objects, it can improve the \emph{F1} score by nearly 3.66\% over the baseline. The AFM is added after the last layer of feature extraction on the basis of CISA to effectively fuse the deep semantic features, which further improves the \emph{F1} score by 3.49\%. The ablation experiments fully demonstrate the effectiveness of our proposed module of CISA, SAEM and AFM.

We validated the effectiveness of SAEM by visualizing the features in Fig. 4. From the visualization of the feature map, we can see that SAEM locates the change features more accurately and obtains a more refined change feature.
\begin{table}[htbp]
\vspace{-3mm}
\renewcommand\arraystretch{1.45}
 \tabcolsep=0.15cm
\renewcommand\arraystretch{1.1}
	\centering
	\caption{Ablation experiments of our proposed module on the LEVIR-CD dataset.}
\begin{tabular}{cclclcclcl}
\hline
\multirow{2}{*}{Methods} & \multicolumn{9}{c}{LEVIR\_CD}                                                                                                                          \\ \cline{2-10} 
                         & \multicolumn{2}{l}{\emph{Pre} (\%)} & \multicolumn{2}{l}{\emph{Rec}(\%)} & \multicolumn{1}{l}{\emph{F1}(\%)} & \multicolumn{2}{l}{FLOPs(G)} & \multicolumn{2}{l}{Params(M)} \\ \cline{1-1}\hline
Base                     & \multicolumn{2}{c}{89.40}    & \multicolumn{2}{c}{85.78}   & 87.55                      & \multicolumn{2}{c}{18.04}    & \multicolumn{2}{c}{\textbf{7.76}}\\
Base+CISA                & \multicolumn{2}{c}{90.78}    & \multicolumn{2}{c}{89.90}   & 90.34                      & \multicolumn{2}{c}{\textbf{7.08}}     & \multicolumn{2}{c}{14.94}     \\
Base+CISA+SAEM           & \multicolumn{2}{c}{91.02}    & \multicolumn{2}{c}{91.40}   & 91.21                      & \multicolumn{2}{c}{7.73}     & \multicolumn{2}{c}{15.95}     \\
Base+CISA+AFM            & \multicolumn{2}{c}{91.21}    & \multicolumn{2}{c}{90.86}   & 91.04                      & \multicolumn{2}{c}{7.19}     & \multicolumn{2}{c}{16.27}     \\
LSAT                     & \multicolumn{2}{c}{\textbf{91.81}}    & \multicolumn{2}{c}{\textbf{91.24}}   & \textbf{91.52}                     & \multicolumn{2}{c}{7.79}     & \multicolumn{2}{c}{16.91}     \\ \hline
	\end{tabular}
	\label{tab3}
\end{table}\vspace{-2mm}
\subsection{Model Efficiency}
The purpose of this Letter is to reduce the computational complexity of Transformer-based CD networks and achieve high detection accuracy. We analyzed and compared the results in terms of floating-point operations (FLOPs), number of parameters (Params), and F1-score (\emph{F1}), which were given in Table IV. We can see that the FLOPs of the proposed LSAT are still smaller than those of most comparison networks, and the detection accuracy \emph{F1} is the best, which can verify the effectiveness of the proposed method. To enable a fair comparison, we replaced the PVT-v2 used in LSAT for ResNet18 in BIT, and we can see that LSAT+Res has a smaller number of parameters and computational costs compared to the lightweight BIT, along with better detection accuracy.\begin{figure*}
\vspace{-3mm}
\setlength{\abovecaptionskip}{-0.3cm}
\setlength{\belowcaptionskip}{-0.cm}
	\begin{center}
		\includegraphics[width=\linewidth]{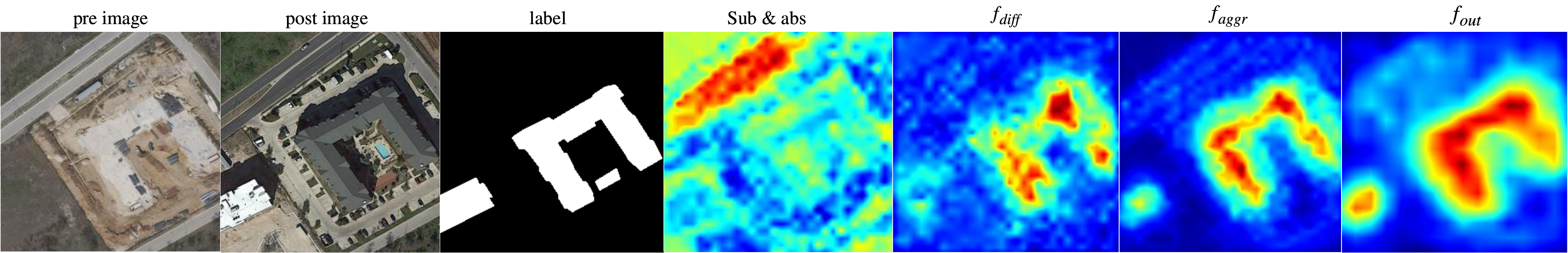}
		\renewcommand{\figurename}{Fig.}
	\end{center}\caption{\small{ Feature visualization. It is clear that SAEM provides better feature maps than vanilla difference operation.}}\vspace{-2mm}
	\label{fig:4}
\end{figure*}

\begin{table}[htbp]
\vspace{-2.5mm}
\renewcommand\arraystretch{1.1}
	\centering
	\caption{Comparative results of model efficiency on the LEVIR-CD dataset.}
\begin{tabular}{cccccc}
\hline
Method Type                  & Network      &\emph{F1}(\%) & FLOPs(G) & Params(M) \\\hline
\multirow{6}{*}{CNN}         & FCN-PP       & 84.64  & 34.65    & 28.13      \\
                             & STANet       & 87.73  & \textbf{6.58}     & 16.93     \\
                             & IFNet          & 87.80  & 41.18    & 50.71     \\
                             & FDCNN        & 85.76  & 32.40    & 13.71     \\
                             & SNUNet       & 88.29  & 33.04    & 12.03     \\
                             & DSAMNet      & 85.48  & 75.29    & 16.95     \\\hline
\multirow{4}{*}{Transformer} & BIT          & 89.31  & 8.44     & 6.93      \\
                             & ChangeFormer & 90.40  & 202.83   & 41.01     \\
                             & SwinSunet    & 90.11  & 11.19    & 40.95     \\
                             & \textbf{LSAT (ours)}  & \textbf{91.52}  & 7.79     & 16.91     \\
                             & LSAT+Res     & 90.21  & 6.64     & \textbf{6.34}      \\\hline
	\end{tabular}\vspace{-3mm}
	\label{tab4}
\end{table}

\section{CONCLUSION}
In this Letter, we have proposed a lightweight structure-aware network LSAT for the VHR RS image CD. The proposed LSAT employs a cross-channel interactive self-attention (CISA) module with linear complexity, which solves the problem of quadratic complexity in self-attention mechanism. In addition, fine-grained change information is obtained using an effective Structure Awareness Enhancement Module (SAEM) and Attention-based fusion module (AFM). Experiments on the publicly available large remote sensing image change detection datasets LEVIR-CD and CDD fully demonstrate the effectiveness of the proposed LSAT.

\end{document}